\documentclass[letterpaper, 10 pt, conference]{ieeeconf}  

\IEEEoverridecommandlockouts                              

\usepackage{times} 
\usepackage{amsmath} 
\usepackage{amssymb}  
\usepackage{algpseudocode,algorithm,algorithmicx}
\usepackage{float}
\usepackage{subfig}
\usepackage{bm}
\usepackage[keeplastbox]{flushend} 
\usepackage[font=small,labelfont=bf]{caption}
\usepackage{graphicx}
\usepackage{multirow}
\usepackage{bm}
\usepackage{booktabs}
\usepackage{array}
\usepackage{gensymb} 
\usepackage{threeparttable}
\usepackage{algpseudocode,algorithm,algorithmicx}
\usepackage[dvipsnames]{xcolor}
\usepackage{cite}
\usepackage{bbm}
\usepackage{dsfont}
\usepackage{mathrsfs}
\usepackage{mathtools}
\definecolor{red}{rgb}{1, 0, 0}
\definecolor{green}{rgb}{0, 1, 0}
\definecolor{grey}{rgb}{0.8,0.8,0.8}
\definecolor{yellow}{rgb}{1,1,0}
\newcolumntype{C}[1]{>{\centering}m{#1}}
\algnewcommand\True{\textbf{true}\space}

\newcommand{\norm}[1]{\left\lVert{#1}\right\rVert}

\title{\LARGE \bf
Frontier-based Automatic-differentiable Information Gain Measure for Robotic Exploration of Unknown 3D Environments
}

\author{Di Deng, Zhefan Xu, Wenbo Zhao, and Kenji Shimada
\thanks{Department of Mechanical Engineering, Carnegie Mellon University, 5000 Forbes Ave, Pittsburgh, PA 15213, USA.,
{\tt\small dengd@andrew.cmu.edu}. 
}
\thanks{The authors would like to thank TOPRISE Co., LTD. for their partial financial support for this work.}
}

\begin{document}
\maketitle
\thispagestyle{empty}
\pagestyle{empty}

\noindent \begin{abstract}
The path planning problem for autonomous exploration of an unknown region by a robotic agent typically employs frontier-based or information-theoretic heuristics. Frontier-based heuristics typically evaluate the information gain of a viewpoint by the number of visible frontier voxels, which is a discrete measure that can only be optimized by sampling. On the other hand, information-theoretic heuristics compute information gain as the mutual information between the map and the sensor's measurement. Although the gradient of such measures can be computed, the computation involves costly numerical differentiation. In this work, we add a novel fuzzy logic filter in the counting of visible frontier voxels surrounding a viewpoint, which allows the gradient of the information gain with respect to the viewpoint to be efficiently computed using automatic differentiation. This enables us to simultaneously optimize information gain with other differentiable quality measures such as path length. Using multiple simulation environments, we demonstrate that the proposed gradient-based optimization method consistently improves the information gain and other quality measures of exploration paths. 

\end{abstract}

\section{Introduction}
\begin{figure}[t]
\centering
\subfloat[Tunnel construction site]{
\includegraphics[width=0.45\linewidth]{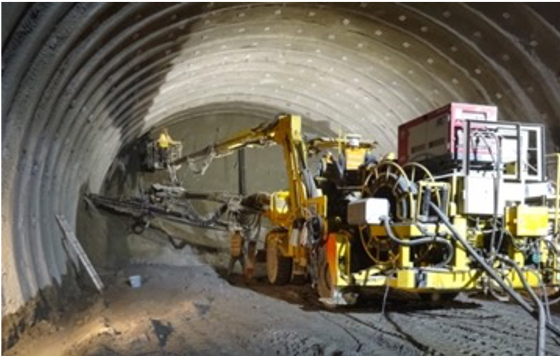}
}   
\subfloat[Simulated virtual environment] {
	\includegraphics[width=0.45\linewidth]{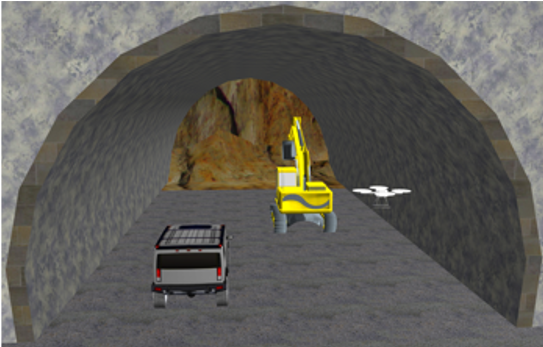}
} \\
\subfloat[Exploration path] {
	\includegraphics[width=0.45\linewidth]{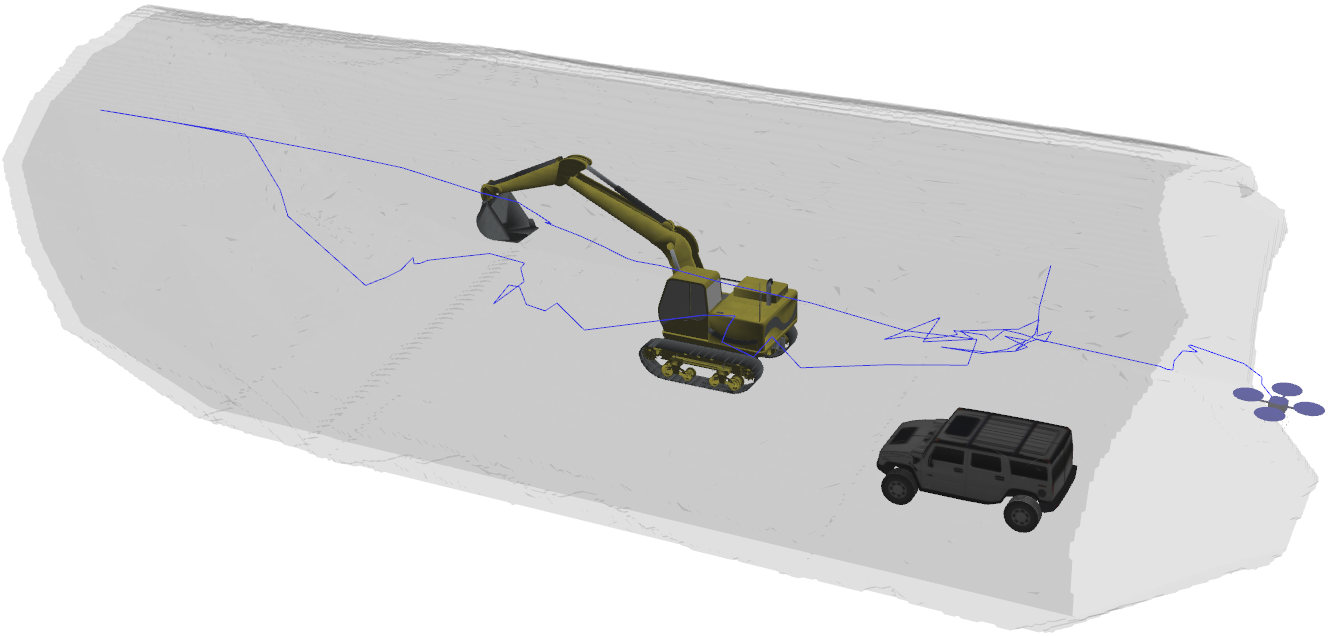}
}
\subfloat[Reconstructed environment] {
	\includegraphics[width=0.45\linewidth]{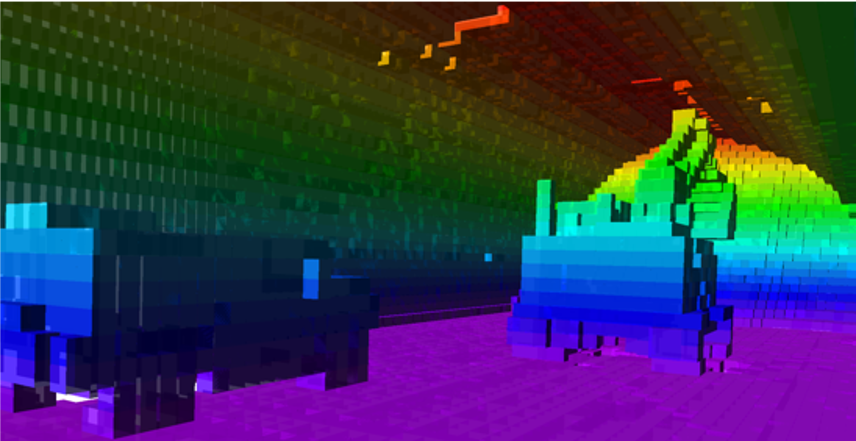}
}
\caption{Path planning for underground tunnel exploration with a RGB-D camera-equipped quadrotor.}
\label{fig:tunnel}
\vspace{-0.4cm}
\end{figure}

\noindent As avionics and on-board sensors get cheaper and more powerful, mobile robots are becoming increasingly ubiquitous in applications such as infrastructure inspection, construction site surveying, urban environment mapping, subterranean cave exploration, livestock surveillance, and underground mine search and rescue. These tasks require robots with navigation autonomy that generate paths completely covering target environments. Fig. \ref{fig:tunnel} shows such exploration tasks, exploration of a tunnel construction site with an autonomous quadrotor. 

One of the main research areas to realize these tasks is the efficient generation of robot paths for exploring unknown 3D environments. As a robot receives sensor measurements, it updates its belief of what the environment looks like. One way to represent this belief is the probabilistic occupancy map \cite{hornung2013octomap}, which classifies subsets of the environment as either free, occupied, or unknown. Next, using this partially reconstructed map, the robot uses some heuristics to compute future paths that are likely to maximize unknown area exploration.

Perhaps the most important part of 3D exploration algorithms is a set of heuristics that encourages the robot to explore unknown space. Existing heuristics can be broadly categorized into: (\textbf{i}) frontier-based \cite{yamauchi1998frontier, holz2010evaluating}, and (\textbf{ii}) information-theoretic \cite{julian2014mutual, bai2016information, amigoni2010information}. Frontier is defined as the boundary between known and unexplored regions of the map. For example, a good measure of the information gain of a viewpoint is the number of visible frontier voxels. Frontier-based heuristics are good at guiding exploration globally, but the their discrete nature makes it difficult to optimize the paths using gradient-based methods. On the other hand, information-theoretic heuristics offer a denser, but more local measure of information gain by calculating the mutual information between the current probabilistic occupancy map and the field of view of the robot's sensor. The gradient of information-theoretic heuristics can be computed, but the computation involves costly numerical differentiation \cite{charrow2015information}.

\textbf{Contribution}. We propose a novel frontier-based information gain measure whose gradient can be efficiently calculated using automatic differentiation \cite{griewank1989automatic}. Unlike mutual information \cite{charrow2015information}, the gradient of the proposed information gain measure can be computed using the same amount of computation as the information gain itself. Together with other terms which achieve objectives such as path length, the information gain of exploration paths is optimized as a non-linear optimization program. Compared with paths generated by sampling-based methods, paths optimized with the proposed method consistently generate higher information gain while reducing the path length.

\section{Related Work}

\noindent \textbf{Frontier-based heuristics}. Frontier is the boundary between free and unknown spaces in an occupancy map. Intuitively, frontier voxels have higher information gain because a robot at the frontier can see more of the unknown regions. Earlier frontier-based method commands the robot to go directly to the frontier voxels \cite{yamauchi1997frontier}. Dornhege \textit{et al.} propose a more sophisticated measure of information gain which considers both frontiers and voids, which are defined as the unknown regions in an occupancy map \cite{dornhege2013frontier}. Shen \textit{et al.} replace the voxel grid representation of free space with a set of particles that behave similarly to gas molecules, and extracts exploration-rewarding goals from particles that have recently reflected from an obstacle \cite{shen2012stochastic}. 

The information gain of a viewpoint can also be evaluated by counting the number of frontier cells visible from the viewpoint. Searching for the viewpoint with the highest information gain is known as the Next Best View (NBV) problem \cite{connolly1985determination}. This search is usually carried out by sampling a large number of viewpoints and counting the number visible frontiers for each of them; the best sample is then used as the goal for the robot \cite{bircher2016three}. Song and Jo extend the counting of visible frontier cells from only the goal to the entire path \cite{song2017online}, but the maximization of visible frontier cells is still achieved by sampling. 

\textbf{Information-theoretic heuristics} reward exploration by finding robot configurations that maximize the mutual information between the current probabilistic occupancy map and the robot's sensor measurements \cite{julian2014mutual}. Unlike the discrete, non-differentiable frontier-based information gain measures, the gradient of mutual information with respect to robot configuration can be computed from finite differences \cite{charrow2015information}, allowing information gain to be used as an objective in trajectory optimization. However, the computational cost of finite differencing is high, and maximizing mutual information can lead to myopic behaviors. As a result, frontier-based measures are often used in conjunction with mutual information as exploration heuristics\cite{charrow2015information}. 

\textbf{Planning a path to the goal}. After finding a goal with the aforementioned exploration heuristics, a path from the robot's current position to the goal can be planned with either a sampling-based algorithm \cite{lavalle1998rapidly} such as Rapidly-exploring Random Tree (RRT) or trajectory optimization \cite{schulman2013finding, ratliff2009chomp}. Sampling-based algorithms are good at avoiding obstacles and local minima \cite{hollinger2014sampling};
optimization-based planning algorithms are better at locally improving the smoothness and safety of the collision-free paths found by sampling-based methods \cite{gao2017gradient}.

\section{Overview of Proposed Planner}
\noindent The main aim of the proposed planning framework is to use a robotic agent (e.g., a quadrotor) equipped with a range sensor (e.g., an RGB-Depth camera) to explore an unknown bounded 3D region, $\textbf{V} \subset \mathbb{R}^3$, in order to determine the subsets of $\textbf{V}$: free ($\textbf{V}_{\text{free}} \subseteq \textbf{V}$), occupied ($\textbf{V}_{\text{occ}} \subseteq \textbf{V}$) and unknown ($\textbf{V}_{\text{unknown}} \subseteq \textbf{V}$). The region, $\textbf{V}$, is typically discretized into small cubes of edge length $\varrho$, where $\varrho$ is the called the resolution of $\textbf{V}$. Each small cube, $\bm{v} \in \textbf{V}$, is called a voxel or a cell.

The occupancy information of $\textbf{V}$ is represented by a probabilistic occupancy map: $\mathbb{M}: \textbf{V} \rightarrow [0,1]$ \cite{thrun2005probabilistic}. For each voxel $\bm{v}$, the occupancy map $\mathbb{M}$ returns the probability of the voxel being occupied. Specifically, $\bm{v}$ is considered free if $\mathbb{M}(\bm{v}) < 0.5$, occupied if $\mathbb{M}(\bm{v}) > 0.5$ and unknown if $\mathbb{M}(\bm{v}) = 0.5$. With every new range sensor measurement, the occupancy probability of every voxel is updated with the binary Bayes filter \cite{moravec1985high}.

We represent $\mathbb{M}$ with Octomap \cite{hornung2013octomap}, a probabilistic occupancy map based on octrees. Octomap allows efficient storage and fast query of the occupancy of any $\bm{v} \in \textbf{V}$, which is essential for viewpoint evaluation and obstacle avoidance.

An overview of the optimization-based exploration framework is given in Alg. \ref{algo:gradient_planning}. Octomap $\mathbb{M}$ of resolution $\varrho$ is initialized in Line 1. Starting at the initial viewpoint, $\bm{\mathsf{q}}_{0}$, the robotic agent scans the surrounding environment with the on-board range sensor, and initializes Octomap $\mathbb{M}$ with its measurements (Line 2). In every exploration iteration, a goal viewpoint, $\bm{\mathsf{q}}_{\text{goal}}$, is first selected as the viewpoint with the highest number of visible frontier voxels from some sampled viewpoints (Line 4). A collision-free path from $\bm{\mathsf{q}}_{0}$ to $\bm{\mathsf{q}}_{\text{goal}}$ is then planned with a sampling-based algorithm such as RRT (Line 5). With the proposed differentiable information gain measure, detailed in Sec. \ref{sec:gradient_based_info_gain_optimization}, the planned path, $\bm{\mathsf{Q}}$, is then optimized to improve its information gain and stiffness (Line 6). Lastly, the robot agent follows the optimized path $\bm{Q}$ and updates the Octomap $\mathbb{M}$ (Lines 7-9). This process repeats until the path information gain becomes less than a threshold, $\epsilon$ (Line 3). The result of the exploration is the probability occupancy map, $\mathbb{M}$, representing the distribution of obstacles and free space in the environment.

\begin{algorithm}
\caption{Gradient-Based Space Coverage}
\label{algo:gradient_planning}
\begin{algorithmic}[1]
\Require $\bm{\mathsf{q}}_{0}$ 
\Ensure $\mathbb{M}$
\State $\mathbb{M}$ = Octomap($\varrho$), IG = $\inf$
\State $\mathbb{M}\gets$ UpdateMap($\bm{\mathsf{q}}_0$, $\mathbb{M}$)
\While{IG $> \epsilon$}
\State $\bm{\mathsf{q}}_{\text{goal}} \gets$ NextBestView($\bm{\mathsf{q}}_0$, $\mathbb{M}$) \;\; // Sec. \ref{subsec:finding_frontier_voxels} to \ref{subsec:penalizing_collision}
\State $\bm{\mathsf{Q}} \gets$ GlobalPlanner($\bm{\mathsf{q}}_{0}$, $\bm{\mathsf{q}}_{\text{goal}}$, $\mathbb{M}$) \quad // Sec. \ref{subsec.next_best_view_and_rrt}
\State $\bm{Q}$, IG $\gets$ GradientPathOptimizer($\bm{\mathsf{Q}}, \mathbb{M}$) \; // Sec. \ref{sec:gradient_based_info_gain_optimization}

\For{$\bm{q}_i$ in $\bm{Q} = \{\bm{q}_1, \ldots \bm{q}_n\}$}
\State $\mathbb{M} \gets$ UpdateMap($\bm{q}_i$, $\mathbb{M}$)
\EndFor
\State $\bm{\mathsf{q}}_0=\bm{\mathsf{q}}_{\text{goal}}$
\EndWhile
\end{algorithmic}
\end{algorithm}

\section{Global Path Planning} \label{sec:global_planner}
\noindent In this section, we present: (\textbf{i}) the computation of a widely used frontier-based information gain measure - the number of visible frontier voxels, and (\textbf{ii}) the sampling-based collision-free path planning algorithm which connects the robot from its current configuration to a goal configuration. 

\subsection{Finding frontier voxels} \label{subsec:finding_frontier_voxels}
\noindent Let us now introduce: (\textbf{i}) a precise definition of the frontier when the space is represented by Octomap, $\mathbb{M}$, and (\textbf{ii}) a method to calculate frontier voxels from $\mathbb{M}$. 

\begin{figure}[htp!]
\centering
\subfloat[Leaf cells at different depths in different color codes.]{
	\includegraphics[width=0.3\linewidth]{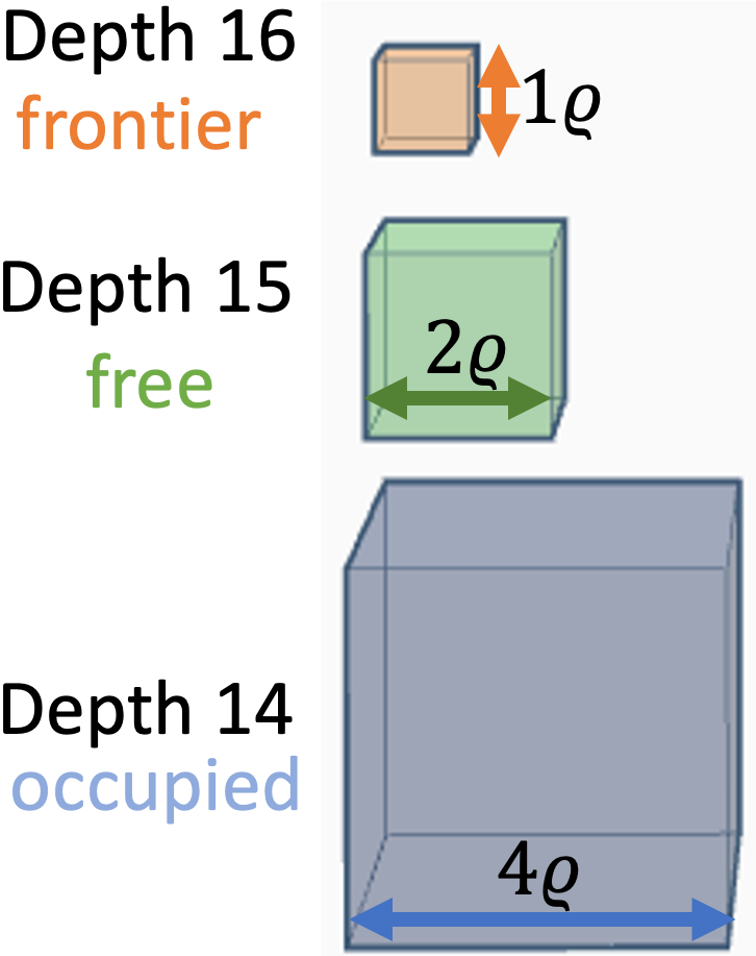}
}
\hspace{0.1cm}
\subfloat[Voxels (shown in orange) surrounding leaf cells (shown in green) at different depths.]{
	\includegraphics[width=0.55\linewidth]{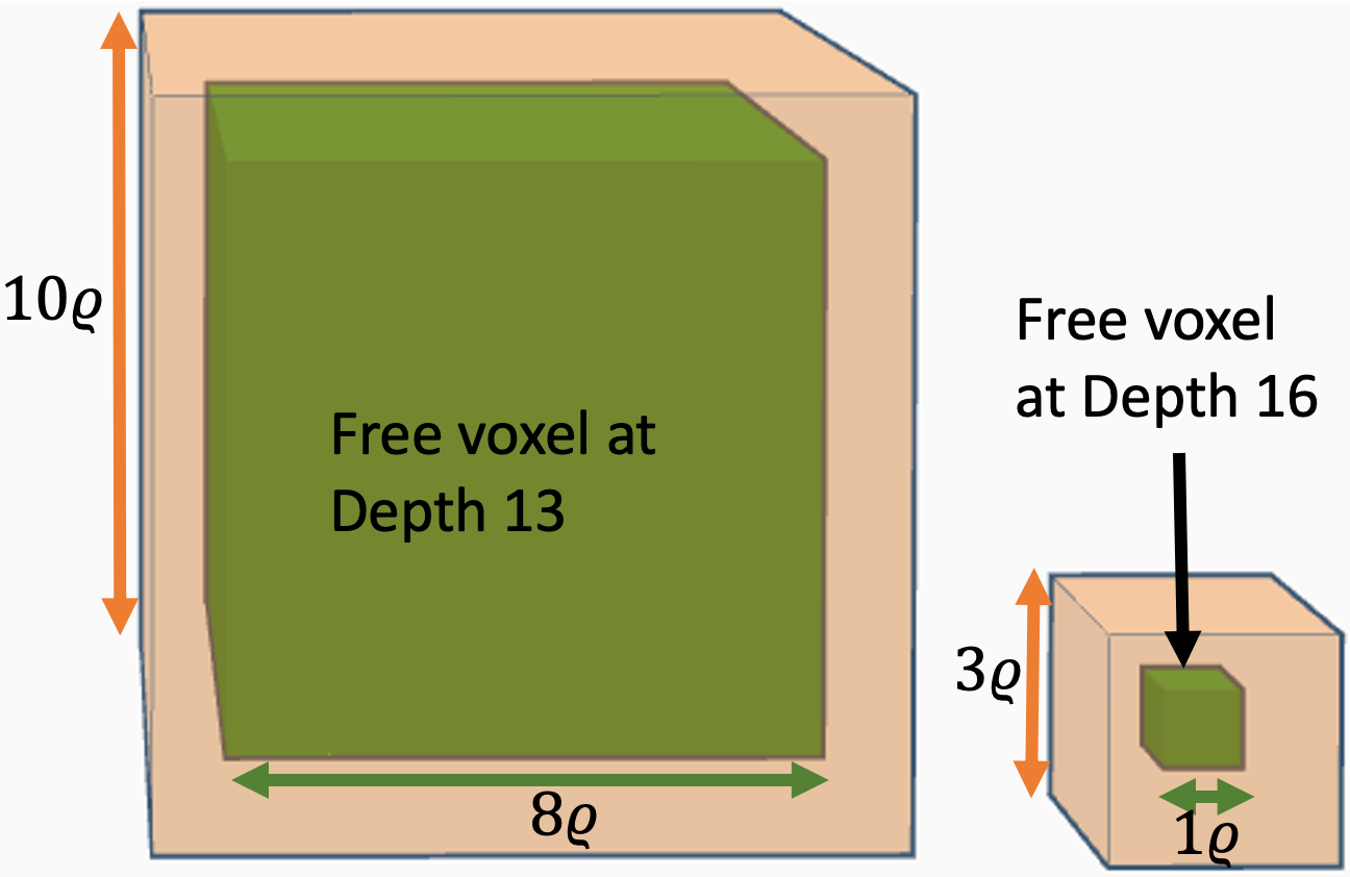}
}
\caption{Region of frontier voxels around free voxels with Depth 13 and 16. $\varrho$ is the minimum size of leaf cells of an Octree.}
\label{fig:octree_depth}
\end{figure}

Octomap fills the explored space, $\textbf{V}_{\text{free}} \cup \textbf{V}_{\text{occ}}$, with leaf cells at different depths in the Octree. The depth represents the rounds of recursive sub-divisions into eight sub-volumes from the entire bounding volume. A leaf cell at the maximum depth (the default is 16 in Octomap) has the same size as a voxel, and the edge length of leaf cells doubles with every decrement in depth, as shown in Fig. \ref{fig:octree_depth}(a).

\begin{figure}[htp!]
    \centering
    \includegraphics[width = 0.4\linewidth]{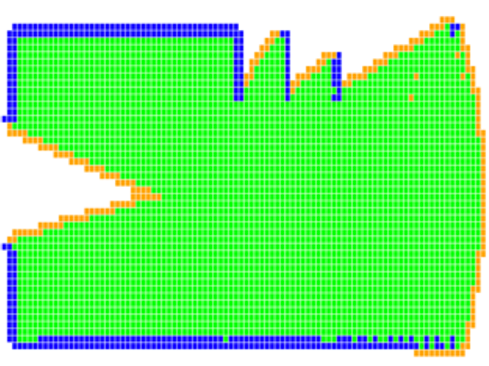}
    \includegraphics[width = 0.4\linewidth]{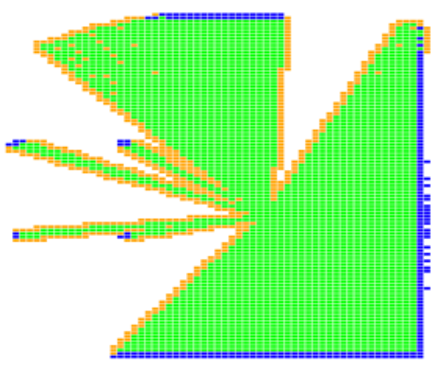}
    \caption{Examples of detected frontier voxels for 2D occupancy grid maps. Free space is shown in green, obstacle in blue, and frontier in orange.}
    \label{fig:frontier}
    \vspace{-0.2cm}
\end{figure}

We store the coordinates of frontier voxels, $(x, y, z)$ in an unordered map. To prevent duplication, we key voxels by the integer cast of $(x, y, z) / \varrho$. Octomap offers an efficient iterator through all leaf cells. For each free leaf cell, i.e., $\mathbb{M}(\text{leaf cell}) < 0.5$, voxels surrounding the leaf cell (Fig. \ref{fig:octree_depth} (b)) are looped through, and voxels which are not yet inside the Octomap are added to the unordered map of frontier voxels. Examples of detected frontier voxels for two different 2D occupancy maps are shown in Fig. \ref{fig:frontier}. 

\subsection{Frontier Voxels Visible to a Camera}
\begin{figure}[htp!]
    \centering
    \includegraphics[width = 0.6\linewidth]{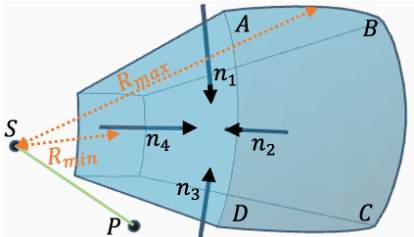}
    \caption{Find Visible Frontiers. Point \textbf{C} is the sensor location. The frustum represents the sensor view region. Point \textbf{P} is the center of a frontier voxel. $\textbf{n}_1$, $\textbf{n}_2$, $\textbf{n}_3$, and $\textbf{n}_4$ are the surface normal pointing inwards.}
    \label{fig:frontier_in_camera_view}
    \vspace{-0.3cm}
\end{figure}

\noindent For a camera with pose $\bm{q}$, a frontier voxel is visible if it is inside the camera's frustum, which is a volume defined by the camera's field of view and depth measurement range $[R_{\text{min}}, R_{\text{max}}]$. As shown in Fig. \ref{fig:frontier_in_camera_view}, the frustum is bounded by four planes and two spherical rectangles with radii equal to $R_{\text{min}}$ and $R_{\text{max}}$. In the camera's body frame $\textbf{S}$, the coordinates of voxel $\bm{v}$, which is originally expressed in world frame $\textbf{W}$, is given by ${}^{\textbf{S}}\bm{v} = \bm{T}^{\textbf{S}}_{\textbf{W}} \bm{v}$, where $\bm{T}^{\textbf{S}}_{\textbf{W}}$ is the homogeneous transformation from $\textbf{S}$ to $\textbf{W}$. The inward normals of the four planes are denoted by $\bm{n}_{1}$ to $\bm{n}_{4}$. Using the above notations, the set of voxels inside the frustum of a camera at $\bm{q}$ is the following set:
\begin{equation}
F(\bm{q}) = \left\{ \bm{v} \in \textbf{V} | \langle \bm{n}_i, {}^\textbf{S}\bm{v} \rangle > 0, \forall i; R_{\text{min}}\le \norm{{}^\textbf{S}\bm{v}} \leq R_{\text{max}} \right\}.
\end{equation}

Even inside the frustum $F$, a voxel $\bm{v}$ can still be invisible to the camera if it is occluded by obstacles. In terms of occupancy probabilities, the set of voxels unobstructed from a camera at $\bm{q}$ is: 
\begin{equation}
\label{eqn.ray}
B(\bm{q}) = \{\bm{v} \in \mathbf{V} \; | \; \mathbb{M}(\bm{v}') < 0.5, \; \forall \bm{v}' \in \textbf{V}_{{}^{\textbf{S}}\bm{v}} \},
\end{equation}
where $\textbf{V}_{{}^{\textbf{S}}\bm{v}}$ is the set of voxels on the line segment between ${}^{\textbf{S}}\bm{v}$ and the camera frame origin, excluding $\bm{v}$ itself. The total number of frontier voxels visible from a camera at pose $q$ can therefore be expressed as:
\begin{equation}
\label{eqn.num_frontier_voxels}
f(\bm{q}) = \sum_{\bm{v} \in \textbf{V}_{\text{frontier}}} \mathds{1}_{F(\bm{q})\bigcap B(\bm{q})}(\bm{v}),
\end{equation}
where $\mathds{1}_{A}(x)$ is the indicator function for set $A$; $\textbf{V}_{\text{frontier}}$ is the set of frontier voxels as identified in Sec. \ref{subsec:finding_frontier_voxels}.

{}

\subsection{Penalizing Collisions}
\label{subsec:penalizing_collision}
\noindent A good viewpoint not only has a larger number of visible frontier voxels, it also needs to stay away from collision. In this section, we illustrate how collision is penalized in the evaluation of the quality of viewpoints. 
\begin{figure}[htp!]
    \centering
    \includegraphics[width = 0.95\linewidth]{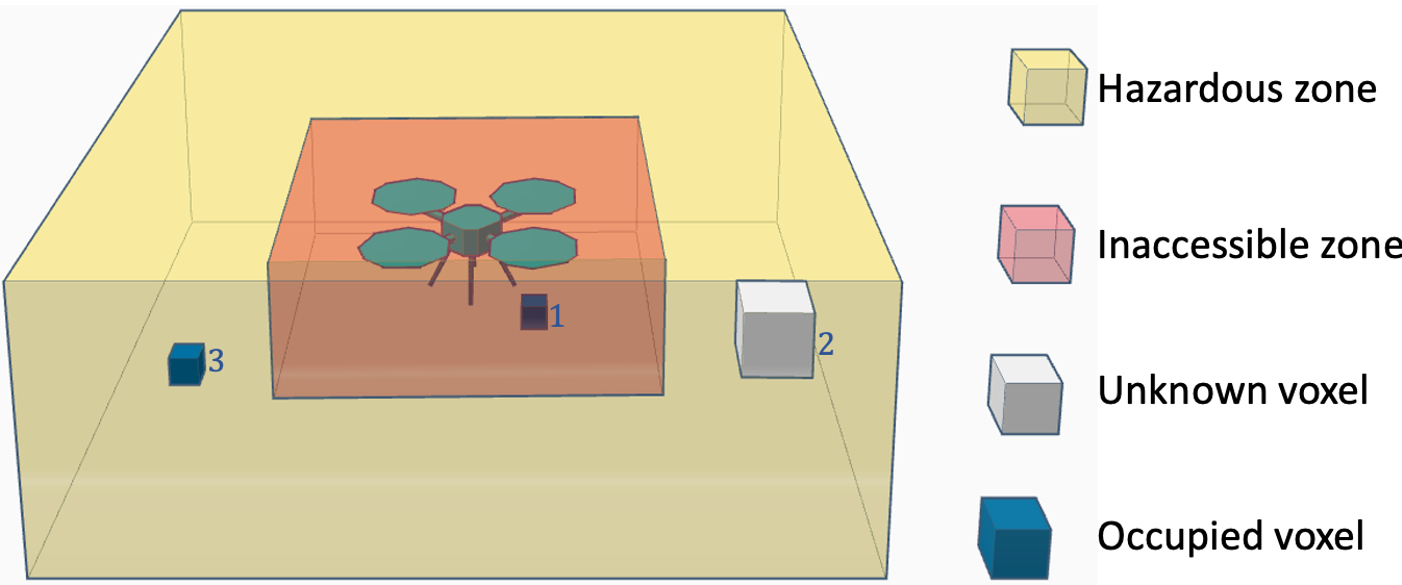}
    \caption{Definition of inaccessible and hazardous zones. $\alpha_1 = 0$ as Obstacle 1 is inside $\textbf{V}_{\text{red}}$. There are two voxels which are unknown or occupied in $\textbf{V}_{\text{yel}}$, thus $\alpha_2 = e^{-3\lambda_2}$.}
    \label{fig:quad_collison}
\end{figure}
We define two bounding boxes around the robotic agent: a smaller inaccessible zone, $\textbf{V}_{\text{red}}$, and a larger hazardous zone, $\textbf{V}_{\text{yel}}$, as shown in Fig. \ref{fig:quad_collison}. No unknown or occupied voxels are allowed inside $\textbf{V}_{\text{red}}$. As the log likelihood ratio of voxel $\bm{v}$, defined as $l(\bm{v}) = \log\frac{\mathbb{M}(\bm{v})}{1-\mathbb{M}(\bf{v})}$, is non-negative if $\bm{v}$ is unknown or occupied, we can define the first collision penalty factor, $\alpha_1$, as follows:
\begin{equation}
\label{eqn.inacc}
    \alpha_1(\bm{q}) = \mathds{1}_{\leq0}\left(\max_{\bf{v}\in\bf{V}_{\text{red}}(\bm{q})}l(\bm{v})\right),
\end{equation}
which is equal to zero if there exists an occupied or unknown voxel in $\textbf{V}_{\text{red}}$. 

In contrast, occupied voxels in $\textbf{V}_{\text{yel}}$ are penalized more mildly:
\begin{equation}
\vspace{-0.1cm}
\label{eqn.danger}
    \alpha_2(\bm{q}) = \exp\left(-\lambda_2 \sum_{\bm{v} \in \bf{V}_{\text{yel}}(\bm{q})}\mathds{1}_{\mathbb{M}(\bm{v})\geq\text{0.5}}(\bm{v}) \right), 
\end{equation}
which drops from 1 to 0 as the number of occupied or unknown voxels in $\bf{V}_{\text{yel}}(\bm{q})$ increases. $\lambda_2$ is a positive user-defined weight. 

\subsection{Selecting and Reaching the Goal}
\label{subsec.next_best_view_and_rrt}
\noindent Firstly, the global path planner samples a large number of viewpoints inside $\textbf{V}_{\text{free}}$ and evaluates their quality using:
\begin{equation}
\label{eqn.IG}
    \text{ViewQuality}(\bm{q}) =  f(\bm{q})\prod_{i\in\{1,2,3\}}\alpha_i(\bm{q}),
\end{equation}
where 
\begin{equation}
\vspace{-0.2cm}
\label{eqn.dist}
    \alpha_3 = e^{-\lambda_3\norm{{\bm{q}-\bm{q}_0}}}
\end{equation}
penalizes the distance between the sample, $\bm{q}$, and the robotic agent's current configuration, $\bm{q}_0$. The view quality measure is the number of visible frontier voxels discounted by collision penalties and distance to $\bm{q}_0$.

\begin{figure}[htp!]
    \vspace{-0.2cm}
    \centering
    \includegraphics[width = 0.50\linewidth]{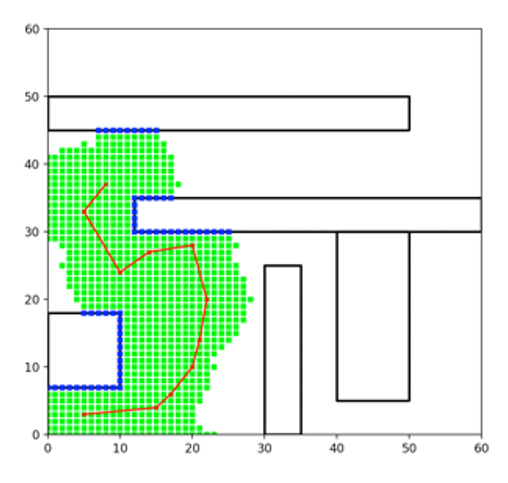}
    \caption{A path planned by RRT in 2D occupancy grid. Green voxels are free and blue are occupied.}
    \label{fig:2d_rrt_path}
    \vspace{-0.3cm}
\end{figure}

The next-best view, or the sample with the highest quality score, is then chosen as goal $\bm{q}_{\text{goal}}$. A path from $\bm{q}_0$ to $\bm{q}_{\text{goal}}$ is then planned using RRT, with configurations sampled from $\textbf{V}_\text{free}$ and collision checks conducted against $\textbf{V}_\text{occ}$. An example of a path planned by RRT in a 2D environment is shown in Fig. \ref{fig:2d_rrt_path}.

\vspace{-0.3cm}
\section{Gradient-based Information Gain Optimization} \label{sec:gradient_based_info_gain_optimization}
\noindent The number of frontier voxels is a discrete information gain measure unsuitable for gradient-based optimization. In this section, we add to the counting of frontier voxels a piecewise linear fuzzy logic filter which makes the information gain function automatic differentiable. We also formulate an optimization program for exploration paths, with the proposed information gain as one of the terms in the objective function.

\subsection{Differentiable information gain for viewpoint $\bm{q}$}
\noindent The number of visible frontier points, as defined in Eqn. (\ref{eqn.num_frontier_voxels}), is a good measure of information gain. However, it is a discrete function and hence hard to optimize using a gradient-based method. 

For a viewpoint, $\bm{q}$, the fuzzy logic filter is a piece-wise linear function that maps a frontier voxel, $\bm{v} \in \textbf{V}_\text{frontier}$, to a value between 0 and 1: $\Phi\left(\bm{q}, \bm{v}\right): \text{SE}(3) \times \textbf{V} \rightarrow [0, 1]$. Intuitively, $\Phi\left(\bm{q}, \bm{v}\right)$ measures how ``close" voxel $\bm{v}$ is from the frustum of a camera located at viewpoint $\bm{q}$. Specifically, we define two types of fuzzy logic: one for translation and one for rotation, which are denoted respectively by $\phi_d$ and $\phi_\theta$.

\begin{figure}[htp!]
    \vspace{-0.2cm}
    \centering
    \includegraphics[width = 0.9\linewidth]{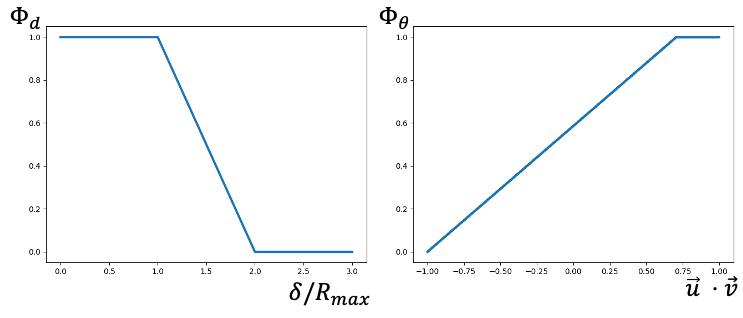}
    \caption{Plots of fuzzy logic filter $\phi_d$ (left) and $\phi_\theta$ (right).}
    \label{fig:fuzzy_logic_filter_plots}
    \vspace{-0.2cm}
\end{figure}

The translational fuzzy logic filter, $\phi_d$, is defined as:
\begin{equation}
    \phi_d(\bm{q}, \bm{v}) = \begin{cases}
  0 &  \delta > 2R_{\text{max}}\\    
  2 - \delta/ R_{\text{max}} & R_{\text{max}}\le \delta \le 2R_{\text{max}} \\
  1  &  \delta < R_{\text{max}}
\end{cases},
\end{equation}
where $\delta = \norm{{}^{\textbf{S}}\bm{v}}$. For a fixed camera pose, $\bm{q}$, the plot of $\phi_d$ as a function of $\delta / R_\text{max}$ is shown in Fig. \ref{fig:fuzzy_logic_filter_plots}. $\phi_d(\bm{q}, \bm{v})$ is flat if voxel $\bm{v}$ inside the ball of radius, $R_\text{max}$, centered at $\bm{q}$, and decreases linearly from 1 to 0 as $\bm{v}$ gets further away from $\bm{q}$. 

The rotational fuzzy logic filter, $\phi_\theta$, is defined as: 
\begin{equation}
\phi_\theta(\bm{q}, \bm{v}) = \begin{cases}
\frac{1+ \langle \bm{u}, \bm{w}\rangle}{1+\cos(\omega / 2)} & \langle \bm{u}, \bm{w}\rangle < \cos(\omega/{2}) \\
1 &  \langle \bm{u}, \bm{w}\rangle \geq \cos(\omega / 2)
\end{cases},
\end{equation}
where $\bm{u}$ is the unit vector along the z-axis of the camera frame, \textbf{S}, and $\bm{w} = {}^{\textbf{S}}\bm{v} / \norm{{}^{\textbf{S}}\bm{v}}$ the unit vector along ${}^{\textbf{S}}\bm{v}$. Similar to the translational case, $\phi_\theta$ is flat if $\bm{v}$ is inside the camera's frustum, and decreases linearly to 0 from the edge of the frustum to the opposite direction along the optical axis of the sensor, -$\bf{u}$ (Fig. \ref{fig:fuzzy_logic_filter_plots}). 

\begin{figure}[htp!]
\vspace{0.2cm}
\centering
\subfloat[Fuzzy logic filter bounding cube] {
	\includegraphics[width=0.46\linewidth]{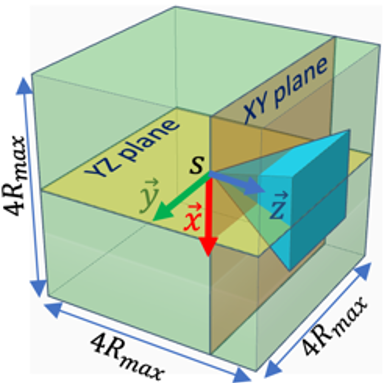}
	\label{fig:discout_factor_a}
}    
\subfloat[YZ plane of the volume] {
	\includegraphics[width=0.46\linewidth]{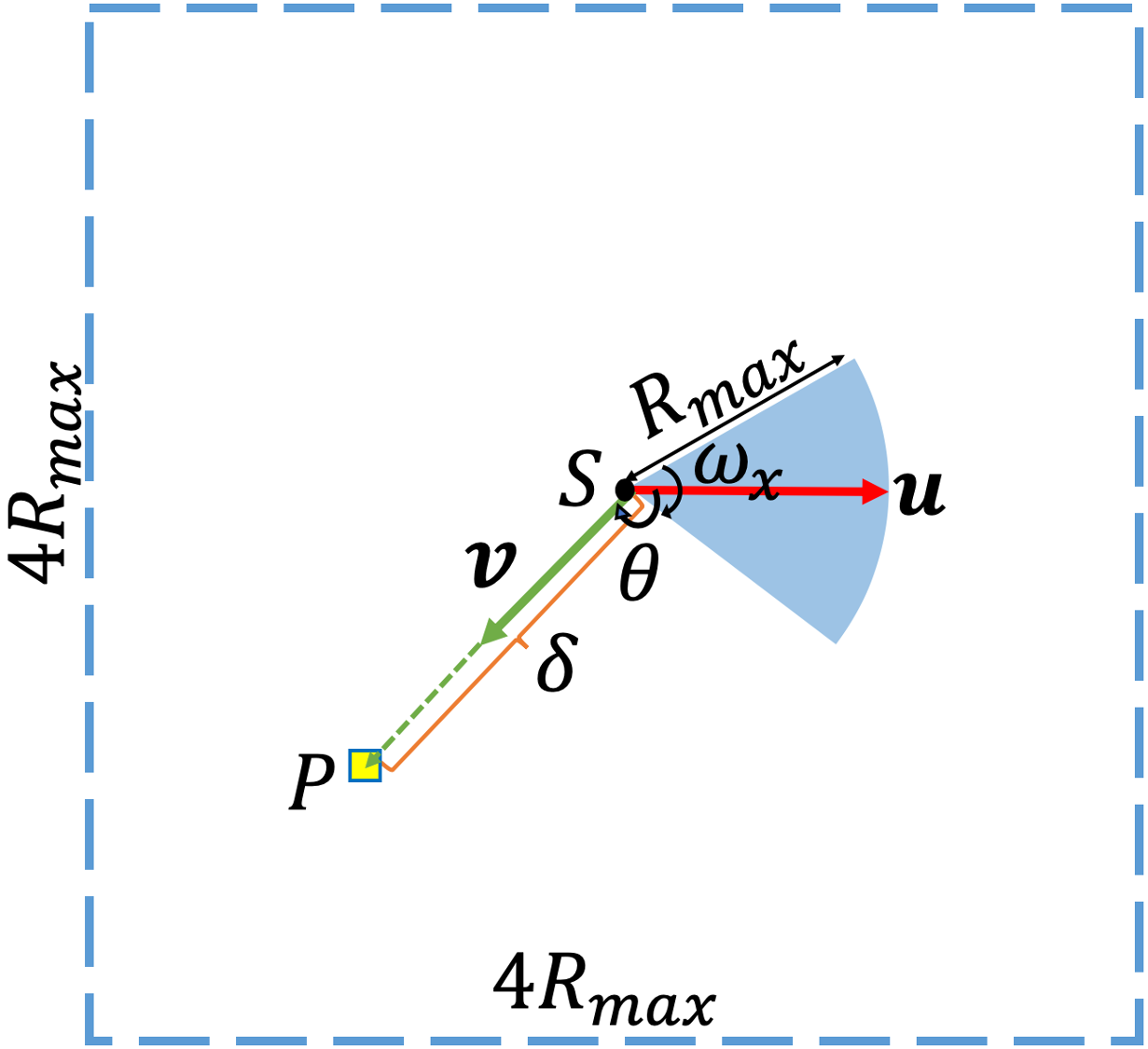}
	\label{fig:discout_factor_b}
}  
\caption{Cubic bounding volume of fuzzy logic filter shown in camera frame $\bf{S}$. $\bm{u}$ is a directional vector of along the optical axis ($\vec{Z}$). The field of view angle of the camera is denoted by $\omega$. $\bf{P}$ is a frontier voxel.}
\label{fig:discout_factor}
    \vspace{-0.4cm}
\end{figure}

Fuzzy logic filter $\Phi$ is therefore defined as the product of the translational and rotational filters:
\begin{equation}
\vspace{-0.1cm}
\Phi(\bm{q}, \bm{v}) = \phi_d(\bm{q}, \bm{v})\phi_{\theta_\text{xz}}(\bm{q}, \bm{v}_\text{xz}) \phi_{\theta_\text{yz}}(\bm{q}, \bm{v}_\text{xz}),
\end{equation}
where $\theta_\text{xz}$ and $\theta_\text{yz}$ denote the rotational fuzzy logic filters in the XZ and YZ plane of the camera frame \textbf{S}, as shown in Fig. \ref{fig:discout_factor_a}. Similarly, $\bm{v}_{\text{xz}}$ and $\bm{v}_{\text{yz}}$ are the projections of ${}^{\textbf{S}}\bm{v}$ onto the XZ and YZ planes.

\begin{figure}[htp!]
\centering
\subfloat[$\phi_{d}$ in the YZ plane defined]{
	\includegraphics[width=0.457\linewidth]{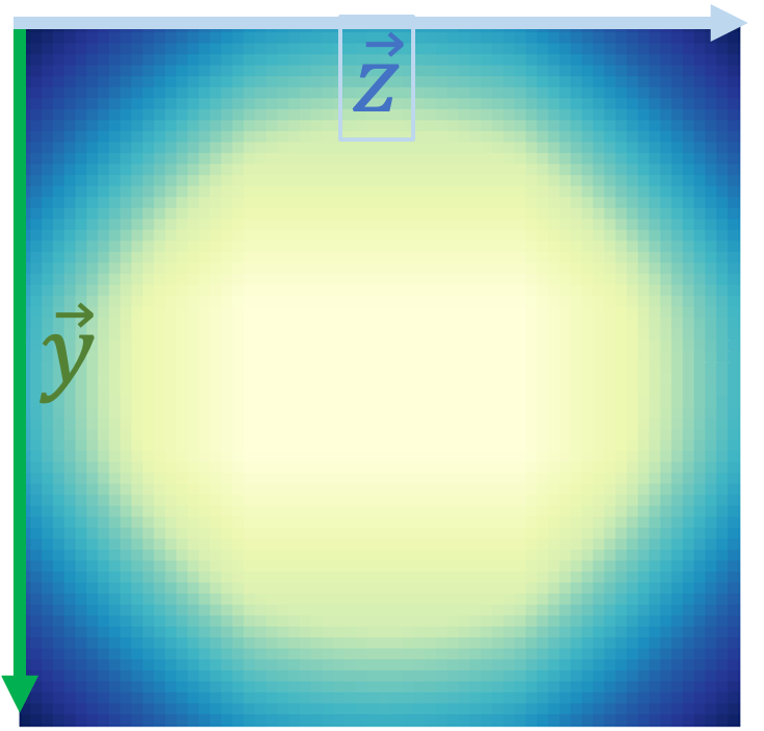}
}    
\subfloat[$\phi_{\theta}$ in the XZ plane]{
	\includegraphics[width=0.473\linewidth]{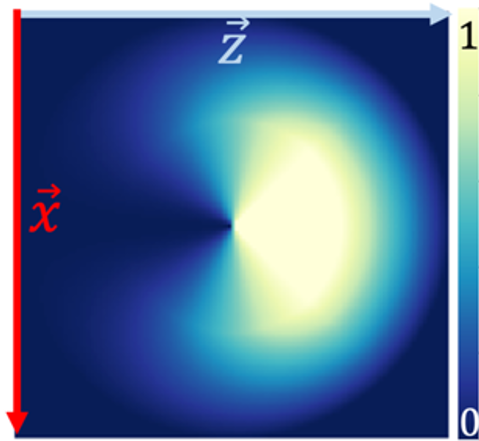}
} 
\caption{Visualization of fuzzy logic filter values in 2D cross-sections defined in Fig. \ref{fig:discout_factor_a}. Voxels located at brighter regions inside the cubic volume have higher values than darker ones.}
\label{fig:discount_map}
\vspace{-0.2cm}
\end{figure}

Note that $\Phi(\bm{q}, \bm{v})$ is differentiable in $\bm{q}$. For a fixed $\bm{v}$, the gradient of $\Phi$ with respect to $\bm{q}$, $\nabla_{\bm{q}} \Phi$, is positive if $\bm{v}$ is outside the frustum of a camera at $\bm{q}$; and incrementing $\bm{q}$ in the direction of $\nabla_{\bm{q}} \Phi$ moves $\bm{v}$ closer to the frustum of the camera. Fig. \ref{fig:discount_map} plots two cross-sections of the fuzzy logic filter $\Phi$ applied in the $(4R_{\text{max}})^3$ volume centered at the origin of \textbf{S}.

We can now define the differentiable information gain of a viewpoint $\bm{q}$ as: 
\begin{equation}
\label{eqn.fuzzy_info}
\text{IG}_{\text{view}}(\bm{q}) = \sum_{\bm{v}\in \bm{V}_{\text{fl}}(\bm{q})} \mathds{1}_{B(\bm{q})}(\bm{v})\Phi(\bm{q}, \bm{v}),
\end{equation}
where $\mathbf{V}_{\text{fl}}(\bm{q})$ represents the set of frontier voxels within a box of edge length $4R_{\text{max}}$ centered at the origin of camera frame \textbf{S} (Fig. \ref{fig:discout_factor_a}). Limiting the calculation of information gain to the box saves redundant calculations on frontier voxels that the camera at $\bm{q}$ is unlikely going to see even after optimization. 

$\text{IG}_{\text{view}}$ is a smoothed version of the number of visible frontier points (Eqn. (\ref{eqn.num_frontier_voxels})). Its gradient can be computed conveniently using automatic differentiation at the same cost as evaluating $\text{IG}_{\text{view}}$ \cite{griewank1989automatic}. The gradient, $\nabla_{\bm{q}} \text{IG}_{\text{view}}$, takes $\bm{q}$ in the direction of increasing the number of visible frontier voxels.

\subsection{Differentiable Information Gain for Path $\bm{Q}$}
\noindent For a path, $\bm{Q} = \{\bm{q}_1, \ldots, \bm{q}_n\}$, consisting of $n$ viewpoints, the calculation of information is almost identical to Eqn. (\ref{eqn.fuzzy_info}) with one exception: a frontier voxel visible from multiple viewpoints in $\bm{Q}$ can only be counted once. 

The function which calculates the path information gain is defined in Algo. \ref{algo:path_info_gain}. Line 3 creates a duplication of the set of frontier voxels, $\bm{V}_{\text{frontier}}$. In Line 4, the first and last configurations in $\bm{Q}$, which correspond to the starting point and goal of the robot, are excluded from the computation of path information gain. This is because the starting point and goal are fixed, making a constant contribution to the total information gain, and hence can be ignored in the optimization of information gain. Line 6 checks whether voxel $\bm{v}$ is inside the $4R_{\text{max}}$ bounding volume and the visibility of $\bm{v}$. Voxel $\bm{v}$ contributes to the information gain only if both checks return true. Furthermore, if $\bm{v}$ is visible from $\bm{q}$ (Line 8), it is erased from $\bf{V}_\text{f}$ (Line 14) so that it does not contribute again to the path information gain.
\begin{algorithm}
\caption{Calculate Information Gain for Path $\bm{Q}$}
\label{algo:path_info_gain}
\begin{algorithmic}[1]
\State \textbf{function} $\text{IG}_{\text{path}}$($\bm{Q}$)
\State $\bf{V}_\text{f} = \bf{V}_{\text{frontier}}$.copy()
\State IG = 0
\For{$\bm{q}$ in $\{\bm{q}_2, \ldots, \bm{q}_{n-1}\}$}
\For{$\bm{v}$ in $\bf{V}_{\text{f}}$}
\If{$\mathds{1}_{\text{BoundingCube}(\bm{q})}(\bm{v})$ and $\mathds{1}_{\bm{B}(\bm{q})}(\bm{v})$}
\State IG += $\Phi(\bm{q}, \bm{v})$
\If{$\mathds{1}_{F(\bm{q})}(\bm{v})$}
\State $\bf{V}_{\text{f}}$.erase($\bm{v}$)
\EndIf
\EndIf
\EndFor
\EndFor
\State \textbf{return} IG
\end{algorithmic}
\end{algorithm}

As $\text{IG}_{\text{path}}$ simply makes repeated calls to $\Phi(\bm{q}, \bm{v})$ and add the returned values, the gradient, $\nabla_{\bm{Q}} \text{IG}_{\text{path}}$, can also be computed using automatic differentiation.

\subsection{Computational Complexity of $\text{IG}_{\text{path}}$}
\noindent The proposed gradient-based path optimization method manages to identify a feasible optimized path within a few seconds to execute onboard for RGB-D cameras within a bounded sensor range selected based on the complexity of $\text{IG}_{\text{path}}$. 

Let $N_q$, $N_f$ and $N$ be the number of viewpoints in a path $Q$, the number of frontier voxels within a fuzzy logic filter bounding cube of a viewpoint and the total number of voxels in the explored map. The complexity of an occupancy query of a voxel is $O(\log N)$. The total number of complexity queries (Line 6 of Algo. \ref{algo:path_info_gain}) is upper-bounded by $(2R_{\text{max}}/\varrho) N_q N_f$. The evaluation of $\Phi(\cdot)$ is fairly trivial ($O(1)$), and the number of calls to $\Phi(\cdot)$ is on the order of $O(N_q N_f)$. Therefore, the worst-case complexity of $\text{IG}_{\text{path}}$ is $O(N_q N_f \log N)$. In practice, the number of frontier voxels whose visibility needs to be checked is usually a lot smaller than $N_f$ as most of them are not inside the bounding cube.

As $\text{IG}_{\text{path}}$ is automatic differentiable, gradient $\nabla_{\bm{Q}} \text{IG}_{\text{path}}$ can be computed at the same complexity as $\text{IG}_{\text{path}}$. In contrast, the complexity of computing $\nabla_{\bm{Q}} \text{IG}_{\text{path}}$ using the finite difference is $\text{IG}_{\text{path}}$ of $O(N_q^2 N_f \log N)$.

\begin{figure*}
\vspace{0.2cm}
\centering
\subfloat[Path by RRT-based global planner and the path's visible frontier voxels.]{
	\includegraphics[width=0.47\textwidth]{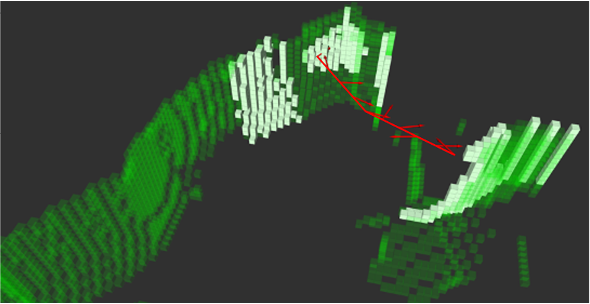}
	\label{fig:frontier_rrt}
}
\subfloat[Path after optimization and the optimized path's visible frontier voxels.]{
	\includegraphics[width=0.47\textwidth]{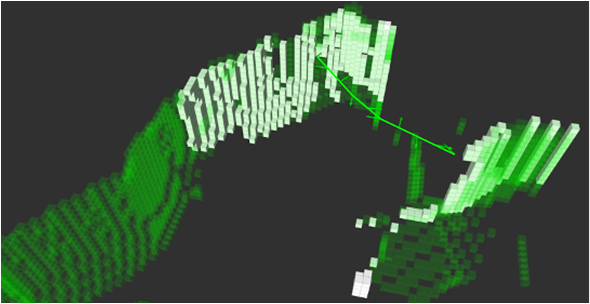}
	\label{fig:fontier_optimization}
}
\caption{Frontier voxels that are visible from paths before (Fig. \ref{fig:frontier_rrt}) and after (Fig. \ref{fig:fontier_optimization}) optimization. The green translucent voxels represent all frontier voxels at the beginning of path index 3 in the exploration of the lab environment (Fig. \ref{fig:env}(a)). The red and green lines represent the paths generated by the global RRT planner and the path after information gain optimization. The white voxels represent frontiers visible to the path planned by their respective planners.}
\label{fig:frontier_comparison}
\vspace{-0.4cm}
\end{figure*}

\subsection{Gradient-based Path Optimization} \label{sec:path_optimization}
\noindent The goal of the path optimization is to improve the initial trajectory returned by the global RRT planner, detailed in Sec. \ref{sec:global_planner}, $\bm{\mathsf{Q}} = \{\bm{\mathsf{q}}_1, \dots, \bm{\mathsf{q}}_k\}$, so that more frontier voxels become visible along the path. 

As the starting point, $\bm{\mathsf{q}}_1$, and the goal point, $\bm{\mathsf{q}}_n$, are fixed, the decision variables of the optimization program are defined as $\bm{Q} = \{\bm{q}_2, \ldots, \bm{q}_{n-1}\}$. In addition to increasing the number of visible frontier points, we also want to penalize the total length of the path. The optimization problem is therefore defined as:
\begin{equation}
\label{eqn:optimization}
    \underset{\bm{Q}}{\text{minimize}} \; \left(- \alpha \text{IG}_{\text{path}}(\bm{Q}) + \beta g_L(\bm{Q}) \right),
\end{equation}
where $\alpha$ and $\beta$ are small regularization constants, and
\begin{align}
    g_L(\bm{Q}) = & \sum_{i=2}^{n-2}\norm{\bm{q}_i - \bm{q}_{i+1}}^2_{\bf{W}} + \\ 
& \norm{\bm{q}_2-\bm{\mathsf{q}}_1}^2_{\bf{W}} + \norm{\bm{\mathsf{q}}_n - \bm{q}_{n-1}}^2_{\bf{W}} 
\label{eqn.obj_dist} 
\end{align}
is the cost on path length. $\norm{\cdot}_{\bf{W}}$ denotes the norm induced by a positive definite matrix, $\bf{W}$.

Optimization problem formulated in Eqn. ($\ref{eqn:optimization}$) is initialized with $\{\bm{\mathsf{q}}_2, \dots, \bm{\mathsf{q}}_{k-1}\}$ and solved using non-linear solvers such as IPOPT \cite{wachter2006implementation}. In practice, we find that primal-dual convergence of Eqn.  ($\ref{eqn:optimization}$) is oftentimes difficult to achieve, but the objective function typically improves and then plateaus after 10-20 iterations. 

Collision avoidance can also be encouraged or enforced by adding the signed-distance function between the robot and the environment to the objective function in Eqn. ($\ref{eqn:optimization}$) or as a constraint. However, the addition of signed distance function slows down computation and makes convergence even more difficult. In practice, we find that it is better to ignore collision at the trajectory planning stage, and to use lower-level motion primitives during execution to avoid imminent collisions.

\section{Results and Discussion}
\noindent In this section, we evaluate the performance of the proposed path optimization method in multiple simulated 3D environments.

\subsection{Simulation design and implementation details}
\noindent The robotic agent used in simulation is a quadrotor with four degrees of freedom: $[x, y, z, \theta]$, where  $[x, y, z]$ are translations in the world frame and $\theta$ is the yaw angle. The quadrotor is equipped with a depth camera with a maximum depth measurement range of 10m and a field of view of $[\frac{\pi}{2},\frac{2\pi}{5}]$ in the XZ and YZ plane, respectively. The inaccessible zone of the quadrotor is 0.6m$\times$0.6m$\times$0.35m and the hazardous zone of the quadrotor is 1.2m$\times$1.2m$\times$0.7m. 

The resolution of the Octomap is set to $\varrho = 0.3\text{m}$. In Eqn. (\ref{eqn:optimization}), we set $\alpha=5\times10^{-4}$ and $\beta = 0.05$.
Weight matrix $\bf{W}$ in Eqn. (\ref{eqn.obj_dist}) is set to 
$\textbf{W} = \text{diag}([1,1,1,0.1])$ so that rotation is penalized less than translation.

\begin{figure*}
\centering
\vspace{0.2cm}
\includegraphics[]{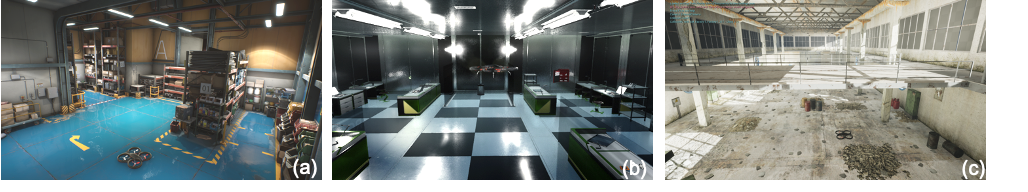}
\caption{Simulation environments. (a) storage room, (b) lab, and (c) factory.}
\label{fig:env}
\end{figure*}

\begin{figure*}
\vspace{-0.2cm}
\centering
\includegraphics[]{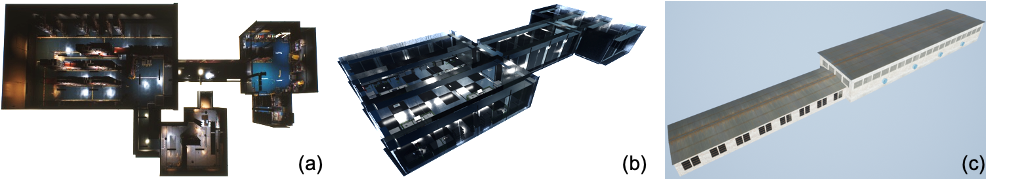}
\caption{Environment geometries. (a) storage room, (b) lab, and (c) factory.}
\label{fig:env_shape}
\end{figure*}

The test environments, including storage room, lab, and factory, are shown in Fig. \ref{fig:env}. Their overall geometries and dimensions are illustrated in Fig. \ref{fig:env_shape} and Table \ref{env_info}. The storage room and factory are large complex environments with narrow passages, numerous obstacles, and partially occluded small regions. Our algorithm is simulated in Airsim \cite{shah2018airsim}.
\begin{table}[t]
\renewcommand\arraystretch{1.2} 

\begin{center}
\vspace{-0.2cm}
\caption{Dimensions of the target environments} \label{env_info}
\begin{tabular}{ |c|c|c|c| } 
 \hline
 Environment & Storage room & Lab & Factory \\
 \hline
 Size ($\text{m}^\text{3}$) 
   & 73$\times$ 32$\times$10 
 & 34$\times$14$\times$4  & 127$\times$17$\times$10 \\  
 \hline
\end{tabular}
\end{center}
\vspace{-0.6cm}
\end{table}


\begin{figure*}
\vspace{-0.2cm}
\centering
\includegraphics[]{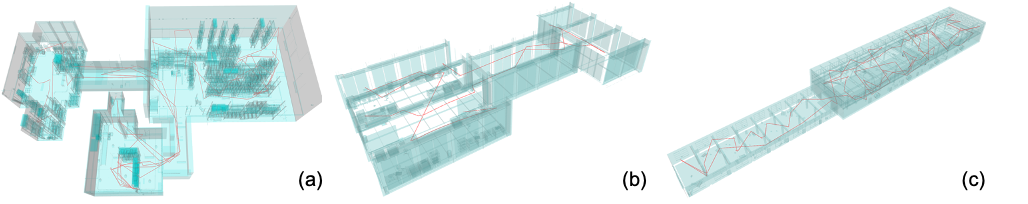}
    \caption{Collision-free path for environment exploration. Red line is the optimized path.}
    \label{fig:path}
\end{figure*}
\begin{figure*}
\centering
\includegraphics[]{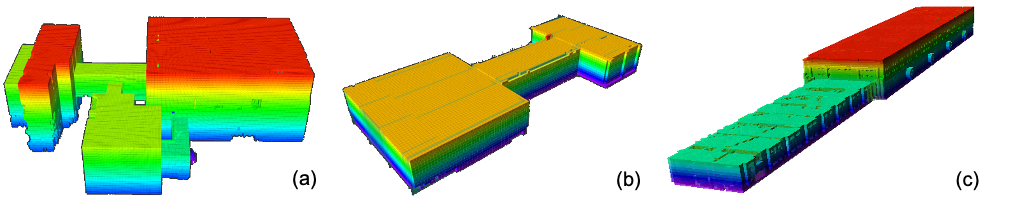}
\caption{Visualization of explored storage room, lab, and factory surfaces: the Octrees are obtained by simulating the planned paths for a quadrotor equipped with RGB-D sensor using the proposed method in Airsim \cite{shah2018airsim}.}
\label{fig:recon}
\end{figure*}

\begin{figure*}
\centering
\includegraphics[width=\linewidth]{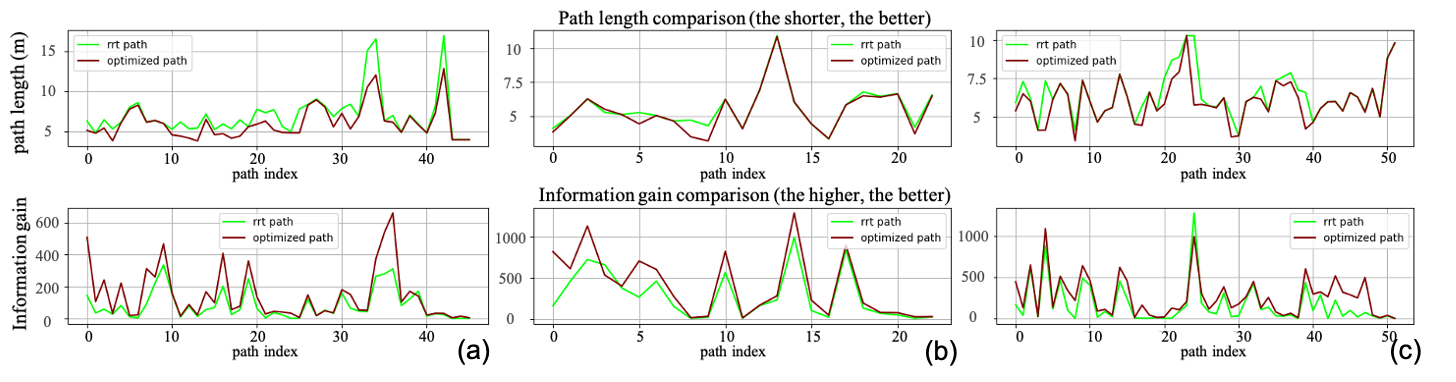}
\caption{Comparison of exploration paths before and after optimization. One increment in path index corresponds to one iteration in the while loop of Alg. \ref{algo:gradient_planning}. The green lines correspond to the paths generated by the RRT-based global planner, and the red lines the paths after optimization.}
\label{fig:compare}
\end{figure*}

\subsection{Path optimization}
\noindent For a single path, $\bm{Q}$, the effect of the path optimization method proposed in Sec. \ref{sec:gradient_based_info_gain_optimization} is shown in Fig. \ref{fig:frontier_comparison}. The number of visible frontier voxels of the optimized path is 10\% higher than the RRT-generated path. In this specific path, we can see the the optimizer is aware of the frontier voxels behind it and nudges the robot to move towards that direction.

\subsection{Exploration}
\noindent The optimized collision-free paths for exploring the storage room, lab, and factory are shown in Fig. \ref{fig:path}, with the corresponding path lengths of 290m, 124m, and 313m. After exploring the storage room for 12.26min, lab for 3.21min, and factory for 11.43min, the reconstructed occupancy grid maps of the environments  shown in Fig. \ref{fig:recon} cover 98.64\% of storage room, 98.91\% of lab, 96.38\% of factory. This performance of our gradient-based path optimization method is compared with the "receding horizon" next-best-view (RH-NBV) algorithm \cite{bircher2016receding} and frontier-based method \cite{yamauchi1997frontier} in the lab and factory environments. The proposed gradient-based optimization method significantly outperforms the other two methods in terms of larger coverage ratio within shorter exploration time in a large complex environment, since it simultaneously improves path length and information gain. Although frontier-based method' coverage ratio is 1\% higher, its path length is 8.8\% longer than our method when exploring the lab. The longer traveling distance is due to the frontier-based method selects viewpoints without evaluating their information gain. Consequently, its coverage ratio drops dramatically as the environment gets more cluttered.

\begin{table}[htp!]
\renewcommand\arraystretch{1.2} 
\begin{center}
\caption{Comparison of the exploration performance in the lab and factory.} \label{optimization_comparison_table}

\begin{tabular}{ |m{1cm} c| m{1cm} m{1cm} m{1.2cm}|}
 \hline
 & Methods  & Coverage ratio & Path Length & Exploration Time \\
 \hline
 \multirow{3}{4em}{\centering {Lab}} & Gradient Optimizer & 98.9\% & \textbf{124.2}m & \textbf{3.21}min\\
 & RH-NBV & 95.8\%  & 140.1m & 6.49min\\
 & Frontier & \textbf{99.9}\% & 135.2m & 5.34min \\
 \hline
 \multirow{3}{4em}{\centering {Factory}} & Gradient Optimizer & \textbf{96.4}\%  & \textbf{313.2}m  & \textbf{11.4}min  \\
 & RH-NBV & 86.5\%  & 419.9m & 36.8min  \\
 & Frontier & 83.8\%  & 369.5m  & 14.9min\\
 \hline
\end{tabular}
\end{center}
\vspace{-0.2cm}
\end{table}

The performance of the storage room, lab, and factory explorations is shown in Fig. \ref{fig:compare}. This figure shows that the paths' information gains never decrease, and most of the time, the optimized paths are smoother than the initial guess after minimizing the objective function. The storage room's exploration path length reduces by 12.51\%, and information gain increases by 73.06\%; the lab's overall path length reduces by 3.83\%, and information gain has grown by 42.78\% after optimization; the overall path length reduces by 5.81\% for the factory, and information gain has increased by 58.02\% after optimization. 

\vspace{-0.1cm}
\section{Conclusion}
\noindent In this paper, we proposed an automated-differentiable information gain measure in 3D indoor environments with a quadrotor equipped with a RGB-D sensor. We are able to improve the information gain of planned paths using gradient-based optimization while reducing the lengths of paths. The effectiveness of our proposed algorithm is verified with three simulated environments.

We believe the performance improvement from the gradient-based optimization can benefit from a more robust measure of information gain, such as clusters of frontier points or information-theoretic measures. We would also like to expand our framework to handle: (\textbf{i}) limited battery life  and transmission distance in large scale environment, and (\textbf{ii}) dynamic constraints of a quadrotor.

\bibliographystyle{IEEEtran}
\bibliography{main}
\end{document}